\documentclass{article}

\usepackage{amsmath}
\usepackage{mathtools}

\newcommand{\leak}{\boldsymbol{\ell}}
\newcommand{\cov}{\mathbf{\Sigma}}

\newcommand{\mv}{\boldsymbol{\mu}_v}

\newcommand{\cc}{\mathbf{c}}
\newcommand{\kk}{\mathbf{k}}
\newcommand{\pp}{\mathbf{p}}

\newcommand{\MM}{\mathbf{M}}

\newcommand{\vv}{\mathbf{v}}
\newcommand{\xx}{\mathbf{x}}
\newcommand{\yy}{\mathbf{y}}
\newcommand{\zz}{\mathbf{z}}

\newcommand{\sdd}{\mathit{SDD}}

\newcommand{\pc}{\mathcal{PC}}

\newcommand{\dat}{\mathcal{D}}
\newcommand{\dattrain}{\mathcal{D}_{\text{train}}}
\newcommand{\dattest}{\mathcal{D}_{\text{test}}}
\newcommand{\datval}{\mathcal{D}_{\text{val}}}

\DeclareMathOperator*{\argmax}{\arg\!\max}

\newcommand{\cbar}{\,|\,}

\usepackage{microtype}
\usepackage{graphicx}
\usepackage{subfigure}
\usepackage{booktabs} % for professional tables
\usepackage[group-separator={,}]{siunitx}

\usepackage{hyperref}

\usepackage[accepted]{icml2024}

\usepackage{amsmath}
\usepackage{amssymb}
\usepackage{mathtools}
\usepackage{amsthm}

\usepackage[capitalize,noabbrev]{cleveref}

\usepackage{tikz}
\usetikzlibrary{bayesnet}
\usetikzlibrary{shapes.symbols}

\usepackage{float}

\theoremstyle{plain}
\newtheorem{theorem}{Theorem}[section]

\theoremstyle{definition}
\newtheorem{definition}[theorem]{Definition}

\theoremstyle{remark}

\usepackage[textsize=tiny]{todonotes}

\icmltitlerunning{Exact Soft Analytical Side-Channel Attacks using Tractable Circuits}

\begin{document}

\twocolumn[
\icmltitle{Exact Soft Analytical Side-Channel Attacks using Tractable Circuits}




\begin{icmlauthorlist}
\icmlauthor{Thomas Wedenig}{tugraz-igi}
\icmlauthor{Rishub Nagpal}{tugraz-iaik}
\icmlauthor{Gaëtan Cassiers}{uclouvain}
\icmlauthor{Stefan Mangard}{tugraz-iaik}
\icmlauthor{Robert Peharz}{tugraz-igi}
\end{icmlauthorlist}

\icmlaffiliation{tugraz-igi}{Institute of Theoretical Computer Science, Graz University of Technology, Graz, Austria}
% \icmlaffiliation{tugraz-igi}{Institute of Machine Learning and Neural Computation, Graz University of Technology, Graz, Austria}
\icmlaffiliation{tugraz-iaik}{Institute of Applied Information Processing and Communications, Graz University of Technology, Graz, Austria}
\icmlaffiliation{uclouvain}{Institute of Information and Communication Technologies, Electronics and Applied Mathematics (ICTM), UCLouvain, Ottignies-Louvain-la-Neuve, Belgium}

\icmlcorrespondingauthor{Thomas Wedenig}{thomas.wedenig@tugraz.at}

\icmlkeywords{Machine Learning, ICML}

\vskip 0.3in
]



\printAffiliationsAndNotice{}  % leave blank if no need to mention equal contribution

\begin{abstract}
Detecting weaknesses in cryptographic algorithms is of utmost importance for designing secure information systems.
The state-of-the-art \emph{soft analytical side-channel attack} (SASCA) uses physical leakage information to make probabilistic predictions about intermediate computations and combines these ``guesses'' with the known algorithmic logic to compute the posterior distribution over the key.
This attack is commonly performed via loopy belief propagation, which, however, lacks guarantees in terms of convergence and inference quality.
In this paper, we develop a fast and exact inference method for SASCA, denoted as ExSASCA, by leveraging knowledge compilation and tractable probabilistic circuits.
When attacking the \emph{Advanced Encryption Standard} (AES), the most widely used encryption algorithm to date, ExSASCA outperforms SASCA by more than $31\%$ top-1 success rate absolute.
By leveraging sparse belief messages, this performance is achieved with little more computational cost than SASCA, and about 3 orders of magnitude less than exact inference via exhaustive enumeration.
Even with dense belief messages, ExSASCA still uses $6$ times less computations than exhaustive inference.

% This document provides a basic paper template and submission guidelines.
% Abstracts must be a single paragraph, ideally between 4--6 sentences long.
% Gross violations will trigger corrections at the camera-ready phase.
\end{abstract}

\section{Introduction}

Unifying learning, logical reasoning and probabilistic inference at scale has a wide range of applications, such as error correcting codes, verification and system diagnostics.
In particular, 
\emph{cryptographic attacks} play a central role in defining, analysing and designing cryptographic algorithms, which form the backbone of our modern information society.
Specifically, \emph{side-channel attacks} attempt to learn about secret material used in a cryptographic computation by observing implementation artifacts, commonly called \emph{leakages}, such as timing information \cite{timing_kocher}, electromagnetic emissions \cite{em_gandolfi}, and power consumption \cite{dpa_kocher}. 
Most contemporary attacks use these leakages to reason about the internal state of a cryptographic algorithm in a probabilistic manner. 
Specifically, in \emph{template attacks} \cite{template_attacks} the attacker has access to a clone of the device under attack and can run the algorithm repeatedly with randomly generated keys and plaintexts as inputs.
Simultaneously, the attacker also records leakage information during all runs, e.g.~the \emph{power trace} of the device. 
Using this data, the attacker can easily build a set of probabilistic models, predicting distributions of intermediate values conditional on the leakage.

The attacker's goal is now to combine these value distributions (``soft guesses'') and the (hard logical) knowledge about the algorithm to infer the posterior distribution of the secret key. 
A na\"ive exhaustive computation of this posterior is prohibitive for practical attacks as a single leakage might require hours of compute time.
As a remedy, the state-of-the-art \emph{soft analytical side-channel attack} (SASCA) \cite{sasca} uses loopy belief propagation (BP) on a factor graph representation of the cryptographic algorithm, to produce an approximate key posterior.
However, loopy BP has limited theoretical underpinnings as neither convergence is granted, nor does it guarantee accurate posterior estimates in case of convergence \cite{bp}. 
Consequently, while SASCA \emph{can} detect weaknesses in cryptosystems, an unsuccessful SASCA does \emph{not} imply any certificate that the leakage cannot be exploited further.
On the contrary, it is straightforward to construct examples where inference is in fact easy but loopy BP fails catastrophically.

In this paper, we develop a fast and exact inference algorithm for SASCA, denoted as \emph{ExSASCA}, for the \emph{Advanced Encryption Standard (AES)}, the most prominent and widely used encryption algorithm to date.
To this end, we build on recent results from knowledge compilation \citep{kcm,sdd} and tractable probabilistic circuits \citep{psdd,vergari2020probabilistic}.
Similarly as in SASCA, we start from a factor graph representation of AES, whose core part, \textsc{MixColumns}, is a highly loopy sub-graph involving $168$ binary variables (bits).
While SASCA performs loopy BP, we instead compile \textsc{MixColumns} into a compact \emph{(probabilistic) sentential decision diagram} (SDD, PSDD) \citep{sdd,psdd}, a circuit representation allowing a wide range of logic and probabilistic inference routines \citep{vergari2020probabilistic}.
On a conceptual level, this amounts to replacing \textsc{MixColumns} with a \emph{single} factor and reducing the factor graph to a \emph{tree}, on which we can perform exact message passing \citep{koller2009probabilistic}, albeit involving a factor with $2^{168}$ entries.
The compiled PSDD is compact, containing only $\num{19000}$ sums and products, and allows inference in \emph{polynomial time of the circuit size}.

In particular, for the key posterior we require mainly \emph{factor multiplication} (worst-case quadratic in the circuit size) and \emph{factor summation} (linear in the circuit size), which are readily provided by tractable circuits and have previously been employed in structured Bayesian networks \citep{sbn} and neuro-symbolic approaches \citep{spl}.
When attacking standard AES, we leverage the fact that the value distributions are typically concentrated on byte values with the same hamming weight, allowing us to work with sparse messages.
With this simplification, an out-of-the box implementation of ExSASCA yields an improvement of \emph{31\% top-1 success rate absolute} in comparison to conventional SASCA.
Here, the required computation for ExSASCA is on the same level as for SASCA, which is \emph{three orders of magnitude} less than for exhaustive enumeration, the only other known exact inference algorithm.

Countermeasures to side-channel attacks, such as \emph{protected} AES implementations, increase the entropy of intermediate algorithmic values and render our sparse-message approach futile. 
For this case, however, we develop a novel dynamic compilation strategy for SDDs, reducing inference to a weighted model counting problem.
The resulting inference machine still requires $6$ times less computation than exhaustive inference and substantially outperforms SASCA on all protection levels.
Overall, our main contributions are:
 
\begin{itemize}
\item We propose ExSASCA, an exact SASCA implementation, which substantially improves the success rate of attacks on AES, while using far less computational resources than inference by exhaustive enumeration.
\item With our method we open a new avenue to study vulnerability in cryptosystems; in the long run our techniques might lead to stronger theoretical guarantees in cryptographic algorithms, for example using results from circuit complexity \citep{de2021compilation} for proving the non-existence of a tractable circuit representation of particular cryptographic algorithms.
\item We develop a novel dynamic compilation framework for circuit-based inference in large-scale-probabilistic systems, which has a wide range of applications, such as error correcting codes \cite{mackay2003information}, system verification and structured Bayesian networks \cite{sbn}.
\end{itemize}

\section{Background}

\begin{figure*}[t]
    \centering
	\includegraphics[width=0.75\linewidth]{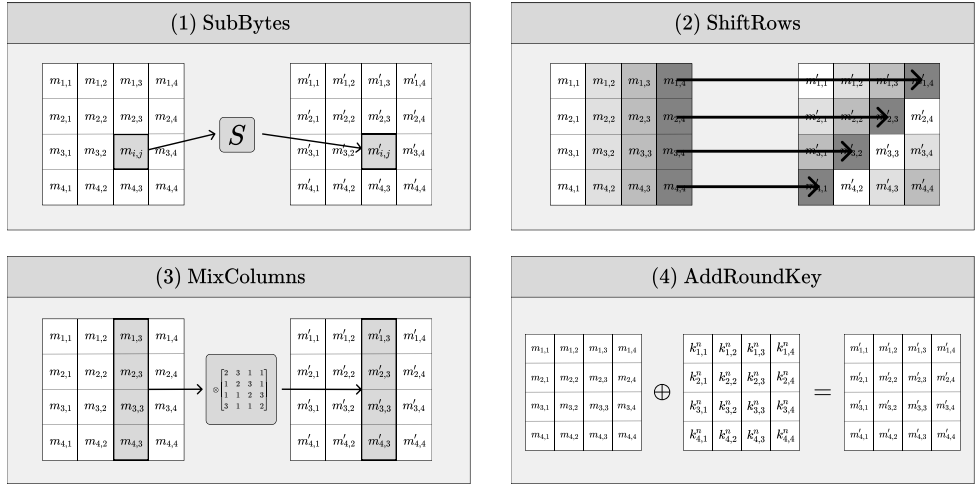}
    \caption{A round of AES takes a $4 \times 4$-byte matrix $\MM$ and computes a series of functions: (1) \textsc{SubBytes} applies a non-linear bijection $S$ to each byte individually, (2) \textsc{ShiftRows} shifts the rows of the input matrix to the left, (3) \textsc{MixColumns} computes a linear function which takes each input column and performs a matrix-vector product in the Galois field of characteristic $256$ ($\mathbb{F}_{256}$) and (4) \textsc{AddRoundKey}, a byte-wise \textsc{xor} operation with the "round key" (which is just the key in the first round).}
    \label{fig:aes_description}
\end{figure*}

\subsection{Advanced Encryption Standard}
In this work, we attack the \textit{Advanced Encryption Standard} (AES) \cite{aes}, the most popular symmetric-key block cipher used to date.
AES takes an $128$-bit \emph{plaintext} $\pp$ and a secret $b$-bit \emph{key} $\kk$ as input, and produces a $128$-bit \emph{ciphertext} $\cc$ as output.
In our experiments, we attack AES-128, i.e., $b=128$.
We refer to bytes
of the key and plaintext as  \emph{input variables} and the bytes of the ciphertext as output variables; besides this, the algorithm also computes several byte-valued \emph{intermediate variables} in the course of computing the ciphertext. 
AES encryption is performed in $10$ iterations (so-called rounds), where each round essentially consists of the functions \textsc{SubBytes}, \textsc{ShiftRows}, \textsc{MixColumns}, and \textsc{AddRoundKey}. 
Figure \ref{fig:aes_description} illustrates such a round.

\subsection{Side-Channel Attacks}

\begin{figure*}[ht]
        \hspace*{3mm}\makebox[\textwidth][c]{
        \scalebox{0.70}{
    \begin{tikzpicture}
    \draw[line width=0.3mm, dashed] (1.9,1) -- (1.9,-8.8);
    \draw[line width=0.3mm, dashed] (3.85,1) -- (3.85,-8.8);
    \draw[line width=0.3mm, dashed] (12.05,1) -- (12.05,-8.8);
    \node[] at (0.4,-8.4) {\small{\textsc{AddRoundKey}}};
    \node[] at (2.9,-8.4) {\small{\textsc{SubBytes}}};
    \node[] at (7.9,-8.4) {\small{\textsc{MixColumns}}};
    
    % 1st row
    \node[latent] (k1) {$k_1$} ; %
    \node[obs, below=0.3cm of k1] (p1) {$p_1$} ; %
    \factor[right=0.5cm of k1, yshift=-0.45cm] {kp1_xor} {below:$\textsc{xor}$} {} {}; %
    \factoredge {k1, p1} {kp1_xor} {} ; %
    \node[latent, right=0.5cm of kp1_xor] (y1) {$y_1$} ; %
    \factor[right=0.5cm of y1] {s1} {below:$\textsc{sbox}$} {} {}; %
    \factoredge {y1} {kp1_xor, s1} {} ; %
    \node[latent, right=0.5cm of s1] (x1) {$x_1$} ; %
    \factoredge {x1} {s1} {} ; %
    \factor[right=0.75cm of x1, yshift=-0.85cm] {xor12} {below:$\textsc{xor}$} {} {}; %
    \node[latent, right=0.5cm of xor12] (x12) {$x_{12}$} ; %
    \factoredge {x12} {xor12} {} ; %
    \factor[right=0.5cm of x12] {xt12} {below:$\textsc{xtime}$} {} {}; %
    \node[latent, right=0.5cm of xt12] (x12_tilde) {$\tilde{x}_{12}$} ; %
    \factoredge {x12, x12_tilde} {xt12} {} ; %
    \factor[right=0.5cm of x12_tilde] {xor12_tilde} {below:$\textsc{xor}$} {} {}; %
    \node[latent, right=0.5cm of xor12_tilde] (x12') {$x_{12}'$} ; %
    \factoredge {x12', x12_tilde} {xor12_tilde} {} ; %
    \factor[right=0.75cm of x12', yshift=0.85cm] {xor1_final} {below:$\textsc{xor}$} {} {}; %
    \node[latent, right=0.5cm of xor1_final] (x1m) {$x_{1}^{(m)}$} ; %
    \factoredge {x1, x12', x1m} {xor1_final} {} ; %
    
    % 2nd row
    \node[latent, below=0.3cm of p1] (k2) {$k_2$} ; %
    \node[obs, below=0.3cm of k2] (p2) {$p_2$} ; %
    \factor[right=0.5cm of k2, yshift=-0.45cm] {kp2_xor} {below:$\textsc{xor}$} {} {}; %
    \factoredge {k2, p2} {kp2_xor} {} ; %
    \node[latent, right=0.5cm of kp2_xor] (y2) {$y_2$} ; %
    \factor[right=0.5cm of y2] {s2} {below:$\textsc{sbox}$} {} {}; %
    \factoredge {y2} {kp2_xor, s2} {} ; %
    \node[latent, right=0.5cm of s2] (x2) {$x_2$} ; %
    \factor[right=0.75cm of x2, yshift=-0.85cm] {xor23} {below:$\textsc{xor}$} {} {}; %
    \factoredge {x2} {s2, xor23} {} ; %
    \node[latent, right=0.5cm of xor23] (x23) {$x_{23}$} ; %
    \factoredge {x23} {xor23} {} ; %
    \factor[right=0.5cm of x23] {xt23} {below:$\textsc{xtime}$} {} {}; %
    \node[latent, right=0.5cm of xt23] (x23_tilde) {$\tilde{x}_{23}$} ; %
    \factoredge {x23, x23_tilde} {xt23} {} ; %
    \factor[right=0.5cm of x23_tilde] {xor23_tilde} {below:$\textsc{xor}$} {} {}; %
    \node[latent, right=0.5cm of xor23_tilde] (x23') {$x_{23}'$} ; %
    \factoredge {x23', x23_tilde} {xor23_tilde} {} ; %
    \factor[right=0.75cm of x23', yshift=0.85cm] {xor2_final} {below:$\textsc{xor}$} {} {}; %
    \node[latent, right=0.5cm of xor2_final] (x2m) {$x_{2}^{(m)}$} ; %
    \factoredge {x2, x23', x2m} {xor2_final} {} ; %
    
    \factoredge {x1, x2} {xor12} {} ; %
    
    % 3rd row
    \node[latent, below=0.3cm of p2] (k3) {$k_3$} ; %
    \node[obs, below=0.3cm of k3] (p3) {$p_3$} ; %
    \factor[right=0.5cm of k3, yshift=-0.45cm] {kp3_xor} {below:$\textsc{xor}$} {} {}; %
    \factoredge {k3, p3} {kp3_xor} {} ; %
    \node[latent, right=0.5cm of kp3_xor] (y3) {$y_3$} ; %
    \factor[right=0.5cm of y3] {s3} {below:$\textsc{sbox}$} {} {}; %
    \factoredge {y3} {kp3_xor, s3} {} ; %
    \node[latent, right=0.5cm of s3] (x3) {$x_3$} ; %
    \factor[right=0.75cm of x3, yshift=-0.85cm] {xor34} {below:$\textsc{xor}$} {} {}; %
    \factoredge {x3} {s3, xor34} {} ; %
    \node[latent, right=0.5cm of xor34] (x34) {$x_{34}$} ; %
    \factoredge {x34} {xor34} {} ; %
    \factor[right=0.5cm of x34] {xt34} {below:$\textsc{xtime}$} {} {}; %
    \node[latent, right=0.5cm of xt34] (x34_tilde) {$\tilde{x}_{34}$} ; %
    \factoredge {x34, x34_tilde} {xt34} {} ; %
    \factor[right=0.5cm of x34_tilde] {xor34_tilde} {below:$\textsc{xor}$} {} {}; %
    \node[latent, right=0.5cm of xor34_tilde] (x34') {$x_{34}'$} ; %
    \factoredge {x34', x34_tilde} {xor34_tilde} {} ; %
    \factor[right=0.75cm of x34', yshift=0.85cm] {xor3_final} {below:$\textsc{xor}$} {} {}; %
    \node[latent, right=0.5cm of xor3_final] (x3m) {$x_{3}^{(m)}$} ; %
    \factoredge {x3, x34', x3m} {xor3_final} {} ; %
    
    \factoredge {x2, x3} {xor23} {} ; %
    
    % 4th row
    \node[latent, below=0.3cm of p3] (k4) {$k_4$} ; %
    \node[obs, below=0.3cm of k4] (p4) {$p_4$} ; %
    \factor[right=0.5cm of k4, yshift=-0.45cm] {kp4_xor} {below:$\textsc{xor}$} {} {}; %
    \factoredge {k4, p4} {kp4_xor} {} ; %
    \node[latent, right=0.5cm of kp4_xor] (y4) {$y_4$} ; %
    \factor[right=0.5cm of y4] {s4} {below:$\textsc{sbox}$} {} {}; %
    \factoredge {y4} {kp4_xor, s4} {} ; %
    \node[latent, right=0.5cm of s4] (x4) {$x_4$} ; %
    \factor[right=0.75cm of x4, yshift=-0.85cm] {xor41} {below:$\textsc{xor}$} {} {}; %
    \factoredge {x4} {s4, xor41} {} ; %
    \node[latent, right=0.5cm of xor41] (x41) {$x_{41}$} ; %
    \factoredge {x41} {xor41} {} ; %
    \factor[right=0.5cm of x41] {xt41} {below:$\textsc{xtime}$} {} {}; %
    \node[latent, right=0.5cm of xt41] (x41_tilde) {$\tilde{x}_{41}$} ; %
    \factoredge {x41, x41_tilde} {xt41} {} ; %
    \factor[right=0.5cm of x41_tilde] {xor41_tilde} {below:$\textsc{xor}$} {} {}; %
    \node[latent, right=0.5cm of xor41_tilde] (x41') {$x_{41}'$} ; %
    \factoredge {x41', x41_tilde} {xor41_tilde} {} ; %
    \factor[right=0.75cm of x41', yshift=0.85cm] {xor4_final} {below:$\textsc{xor}$} {} {}; %
    \node[latent, right=0.5cm of xor4_final] (x4m) {$x_{4}^{(m)}$} ; %
    \factoredge {x4, x41', x4m} {xor4_final} {} ; %
    
    \factoredge {x3, x4} {xor34} {} ; %
    \factoredge {x1, x4} {xor41} {} ; %
    
    % global
    \factor[above=1.25cm of x12, xshift=1.00cm] {global_xor} {above:$\textsc{xor}$} {} {}; %
    \node[latent, right=0.85cm of global_xor] (g) {$g$} ; %
    \factoredge {x12, x34} {global_xor} {} ; %
    \factoredge {g} {global_xor, xor12_tilde, xor23_tilde, xor34_tilde, xor41_tilde} {} ; %
\end{tikzpicture}
        }
       \hspace*{2mm} 
        \scalebox{0.70}{
    \begin{tikzpicture}
    \node[] at (4.5,-4.4) { 
    $\textsc{xor}(a, b, c) =  \begin{cases}
        1 \quad \text{if $a \oplus b = c$} \\
        0 \quad \text{else}
    \end{cases}$ \quad
    $\textsc{sbox}(a, b) =  \begin{cases}
        1 \quad \text{if $S(a) = b$} \\
        0 \quad \text{else}
    \end{cases}$
    };
    \node[] at (4.5,-6.4) { 
    $\textsc{xtime}(a, b)  = \begin{cases}
        1 \quad \text{if $f_{\textsc{xtime}}(a) = b$} \\
        0 \quad \text{else}
    \end{cases}$
    };

    % 1st row
    \node[latent] (k1) {$k_i$} ; %
    \factor[above=0.25cm of k1] {pk1} {above:$p(k_i|\leak)$} {} {}; %
    \factoredge {k1} {pk1} {} ; %
    \node[obs, below=0.3cm of k1] (p1) {$p_i$} ; %
    \factor[right=0.5cm of k1, yshift=-0.45cm] {kp1_xor} {below:$\textsc{xor}$} {} {}; %
    \factoredge {k1, p1} {kp1_xor} {} ; %
    \node[latent, right=0.5cm of kp1_xor] (y1) {$y_i$} ; %
    \factor[right=0.5cm of y1] {s1} {below:$\textsc{sbox}$} {} {}; %
    \factor[above=0.25cm of y1] {py1} {above:$p(y_i|\leak)$} {} {}; %
    \factoredge {y1} {kp1_xor, s1, py1} {} ; %
    \node[latent, right=0.5cm of s1] (x1) {$x_i$} ; %
    \factor[above=0.25cm of x1] {px1} {above:$p(x_i|\leak)$} {} {}; %
    \factoredge {x1} {s1, px1} {} ; %

    \factor[right=1.25cm of x1] {pc} {below:$\mathcal{M}$} {} {}; %

    \node[latent, right=1cm of pc, yshift=1.40cm] (x1m) {$x_i^{(m)}$} ; %
    \factor[right=0.75cm of x1m] {px1m} {above:$p(x_i^{(m)}|\leak)$} {} {}; %
    \factoredge {x1m} {px1m} {} ; %

    \node[latent, below=1.4cm of x1m] (zi) {$v_i$} ; %
    \factor[right=0.75cm of zi] {pzi} {above:$p(v_i|\leak)$} {} {}; %
    \factoredge {zi} {pzi} {} ; %

    \plate {} {(zi)(pzi)(pzi-caption)} {$v_i \in \mathbf{v}_{\mathit{mid}}$} ; %
    \plate {} {(p1)(px1)(px1-caption)(pk1)(pk1-caption)} {$1 \leq i \leq 4$} ; %
    \plate {} {(x1m)(px1m)(px1m-caption)} {$1 \leq i \leq 4$} ; %

    \factoredge {x1, x1m, zi} {pc} {} ; %
\end{tikzpicture}
        }
    }%
    \caption{(left) Factor graph over the first four key bytes and plaintext bytes and their operations in AES.
    Black squares denote logical factors that represent AES operations.
    For example, an \textsc{xor}-factor is $1$ if and only if the variable on its right is the \emph{bit-wise exclusive or} of the two variables on its left and $0$ otherwise.
    Similarly, \textsc{sbox} encodes a  bijection $S$ between byte-values and \textsc{xtime} encodes a multiplication with $2$ in the Galois field of characteristic $256$ (abstracted as $f_{\textsc{xtime}}$).
    Every unobserved (blank) variable node has an additional factor $p(v \cbar \leak)$ (omitted for sake of visual clarity).
    Shaded variable nodes are observed. 
    (right) The same factor graph, but where the loopy \textsc{MixColumns} part has been summarized in a single high-dimensional factor $\mathcal{M}$ (represented by a PSDD). We use plate notation to illustrate structurally identical parts. The set $\vv_{mid}$ contains all intermediate variables in the \textsc{MixColumns} function.}
    \label{fig:sasca}
\end{figure*}
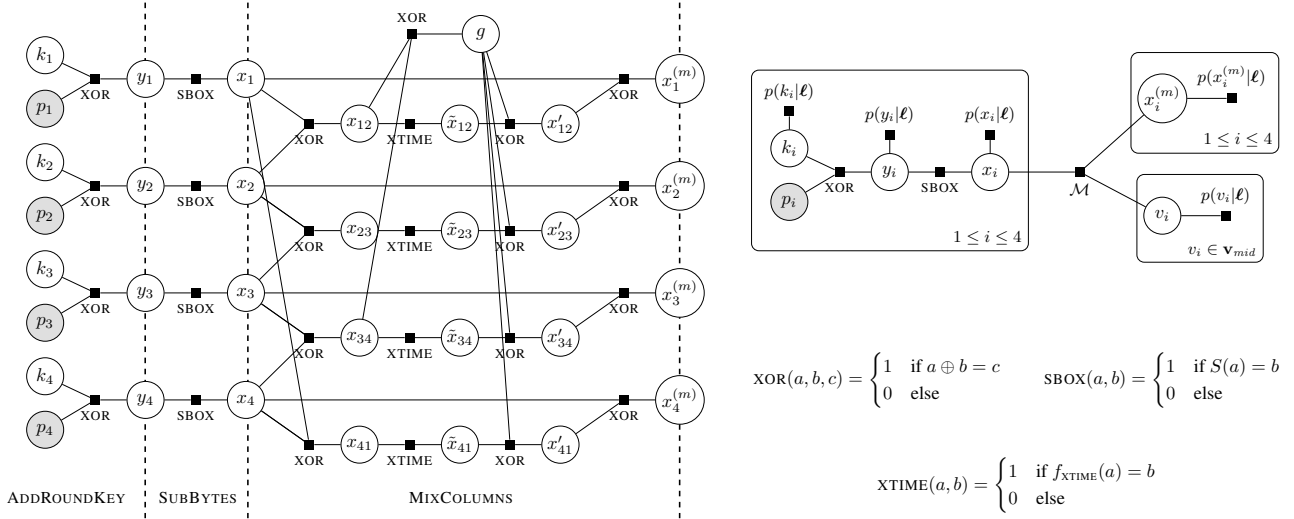

Although a cryptographic algorithm may be secure conceptually, its implementation may be not: Side-channel attacks exploit the fact that physical implementations of an algorithm unintentionally leak information about the processed data \cite{scas}. For example, since the power usage of CMOS transistors depends on the switching activity during the computation, the power consumption of the device performing the encryption is data dependent \cite{intro_sca}. It has been frequently shown that side-channel attacks are more efficient than the best-known cryptanalytic attacks, which consider the system under attack as an idealized, mathematical model \cite{intro_sca}.
While adversaries can exploit various side channels such as timing of operations \cite{timing_kocher} and electromagnetic emanation \cite{em_gandolfi}, this work focuses on power traces. 

\subsubsection{Template Attacks}
\label{sec:ta}

In a template attack the adversarial has access to a \textit{clone} of the target device (e.g., a smart card). 
First, in the so-called \textit{profiling phase}, the attacker queries the clone device with a large number of random inputs, i.e.~key and plaintext in case of AES, and records side-channel leakage of their choice. 
Let $v$ be an arbitrary byte-valued variable involved in AES (input, intermediate variable, or output).
The profiling phase yields a dataset $\{(\leak^{(i)}, v^{(i)})\}_{i=1}^n$, where $n$ is the number of profiling runs and $\leak \in \mathbb{R}^d$ denotes the leakage. 
This data is then used to construct a  likelihood model $p(\leak \cbar v)$, a so-called \emph{template}.
A common choice are multivariate Gaussians with mean $\mv$ and covariance $\cov_v$, i.e., $p(\leak \cbar v) = \mathcal{N}(\leak; \mv, \cov_v)$ for each value $v \in \{0,\dots,255\}$.
Hence, $n$ is typically at least a few thousand, to make sure that for each value there are sufficiently many leakage samples.

Since the leakages $\leak$ are high-dimensional (e.g., $d \approx 10^5$ for power traces), one usually first applies  some dimensionality reduction. 
We adopt the approach from \cite{5min}, using a combination of interest point detection in $\leak$ and {linear discriminant analysis}.
We also experimented extensively with various neural network architectures to construct more expressive templates, which, however, consistently performed worse than \cite{5min}.
Finally, in the attack phase, only the leakage $\leak$ is observed.
Using Bayes' law and a uniform prior $p(v)$, we get a \emph{local distribution} (\emph{belief}) for $v$, conditional on the leakage:
\begin{align}
\label{eq:bayes_simple}
    p(v \cbar \leak) = \frac{p(\leak \cbar v)}{\sum_{v'=0}^{255} p(\leak \cbar v')}
\end{align}
The procedure above, explained for a generic value $v$, is done for several byte variables $v_1, \dots, v_k$ computed in AES.
Usually, one selects variables which are ``close to the key'' in the computational path, as these will typically be more correlated with the secret key than variables ``further away.''

\subsubsection{Soft Analytical Side-Channel Attacks}

After obtaining distributions $p(v_1 \cbar \leak), \dots, p(v_k \cbar \leak)$ for all variables of interest, the attacker aims to aggregate them into a posterior distributions over the \textit{key bytes}. 
Intuitively, the beliefs about intermediate variables, which depend operationally and logically on the secret key, should be propagated backwards and combined into beliefs about the key. 
This is naturally expressed using a \textit{factor graph} \citep{kschischang2001factor, sasca}, where variable nodes (circles) correspond to the byte-valued variables and factors (black squares) represent indicator functions that model the logical relationship between variables.

Figure \ref{fig:sasca} (left) shows the factor graph for part of the first round of AES, consisting of \textsc{AddRoundKey}, \textsc{SubBytes} and one column multiplication of \textsc{MixColumns}. We do not model the \textsc{ShiftRows} operation since it merely corresponds to a fixed re-labeling of the input bytes and does not affect probabilistic inference.
Note that the nodes representing the plaintext are shaded, meaning that they are observed, which is a common assumption in this type of attacks \citep{sasca}.
While the hardware implementation of the \textsc{sbox} also leaks information in practice, we simply abstract it as a $256 \times 256$-dimensional binary factor (i.e., a lookup table).

Every variable (blank node) in the factor graph is equipped with a local distribution derived via a template attack, as described in Section~\ref{sec:ta}.
Note that the figure shows only one of the four parallel branches of the first AES round; the other three branches are of identical structure and can be treated independently.
In this work we attack, as common in literature \cite{sasca}, only the first round of AES; attacking later rounds is doable, but is of diminishing return.

The desired posterior over the key bytes is now simply given as a \emph{factor product} of all involved factors, followed by a summation over all unobserved variables, except for the key bytes \citep{kschischang2001factor}.
Evidently, a na\"ive implementation is infeasible, since the total factor product is of size $2^{232}$ (involving 29 bytes).
However, exploiting the fact that the factor graph encodes an algorithm and that all variables deterministically follow from the key (and plaintext), one actually just needs to enumerate the values of the $4$ key bytes and evaluate the factor graph for each of these $2^{32}$ combinations.
The desired key posterior is then given by re-normalizing the computed factors.

As inference by exhaustive key enumeration is computationally expensive, the \textit{soft analytical side-channel attack} (SASCA) \cite{sasca} applies loopy belief propagation (BP) \citep{kschischang2001factor} to approximate the desired conditional  distribution over key bytes $p(k_i \cbar \leak)$. 
Loopy BP requires about three orders of magnitude less compute than exhaustive inference and is thus considered a practically relevant attack.

However, while SASCA \emph{might} detect weaknesses in a cryptoalgorithm, an unsuccessful application of SASCA is no certificate towards security of the attacked system. 
In particular, loopy BP is notoriously unpredictable, as neither convergence is granted, nor does convergence imply that the posterior has been successfully approximated \citep{bp}.
In fact, it is straightforward to design easy inference problems where loopy BP fails catastrophically.

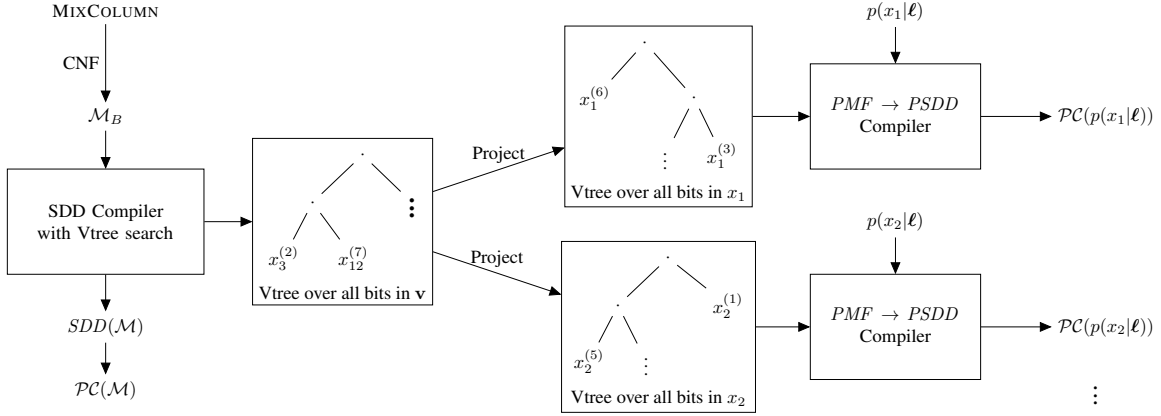
\begin{figure*}[ht]
    % \centering
    \makebox[\textwidth][c]{
        \scalebox{0.70}{
            \begin{tikzpicture}%[trim left=5cm]
    %     \draw (0,0) rectangle node{Test} (2,2);
        \node (m) at (5,4) {$\textsc{MixColumn}$};
        \node (mb) at (5,2) {$\mathcal{M}_B$};
        \node[text width=3.5cm, align=center] (sddcomp) at (5,0) [draw,minimum width=2cm,minimum height=2cm] {SDD Compiler with Vtree search};

        \draw[->] (m) -- (mb) node[midway,left] {CNF};
        \draw[->] (mb) -- (sddcomp) node[midway,above] {};

        \node (sdd_m) at (5,-2) {$\sdd(\mathcal{M})$};
        \node (psdd_m) at (5,-3.2) {$\pc(\mathcal{M})$};
        \draw[->] (sddcomp) -- (sdd_m) node[midway,above] {};
        \draw[->] (sdd_m) -- (psdd_m) node[midway,above] {};
        
        \node[align=center] (vtree_full) at (9.5,0) [draw] {
            \begin{tikzpicture}
            \node[] (dot) {$\cdot$} ; %
            \node[below=0.5cm of dot, xshift=-0.95cm] (a) {$\cdot$} ; %
            \node[right=1.5cm of a] (b) {\huge $\vdots$} ; %
    
            \node[below=0.5cm of a, xshift=-0.55cm] (al) {$x_3^{(2)}$} ; %
            \node[right=0.5cm of al] (bl) {$x_{12}^{(7)}$} ; %
            \edge[-] {a, b} {dot}; %
            \edge[-] {al, bl} {a}; %
            \end{tikzpicture} \\
            Vtree over all bits in $\vv$
        };
        
        \node[align=center] (vtree_x1) at (15.5,2) [draw] {
            \begin{tikzpicture}
            \node[] (dot) {$\cdot$} ; %
            \node[below=0.5cm of dot, xshift=-0.95cm] (a) {$x_1^{(6)}$} ; %
            \node[right=1.3cm of a] (b) {$\cdot$} ; %
    
            \node[below=0.5cm of b, xshift=-0.55cm] (al) {$\vdots$} ; %
            \node[right=0.5cm of al] (bl) {$x_{1}^{(3)}$} ; %
            \edge[-] {a, b} {dot}; %
            \edge[-] {al, bl} {b}; %
            \end{tikzpicture} \\
            Vtree over all bits in $x_1$
        };
        
        \node[align=center] (vtree_x2) at (15.5,-2) [draw] {
            \begin{tikzpicture}
            \node[] (dot) {$\cdot$} ; %
            \node[below=0.5cm of dot, xshift=-0.95cm] (a) {$\cdot$} ; %
            \node[right=1.5cm of a] (b) {$x_2^{(1)}$} ; %
    
            \node[below=0.5cm of a, xshift=-0.55cm] (al) {$x_2^{(5)}$} ; %
            \node[right=0.5cm of al] (bl) {$\vdots$} ; %
            \edge[-] {a, b} {dot}; %
            \edge[-] {al, bl} {a}; %
            \end{tikzpicture} \\
            Vtree over all bits in $x_2$
        };
        
        \node (m) at (23.8,-3.2) {\Large{$\vdots$}};
        
        \draw[->] (sddcomp) -- (vtree_full) node[midway,above] {};
        \draw[->] (vtree_full) -- (vtree_x1) node[midway,above] {Project};
        \draw[->] (vtree_full) -- (vtree_x2) node[midway,above] {Project};
        
        \node[text width=3cm, align=center] (pmfcomp1) at (20,2) [draw,minimum width=2cm,minimum height=2cm] {$\mathit{PMF} \to \mathit{PSDD}$ \\ Compiler};
        \node (px1) at (20,4) {$p(x_1 | \leak)$};
        \draw[->] (vtree_x1) -- (pmfcomp1) node[midway,above] {};
        \draw[->] (px1) -- (pmfcomp1) node[midway,above] {};
        
        \node[text width=3cm, align=center] (pmfcomp2) at (20,-2) [draw,minimum width=2cm,minimum height=2cm] {$\mathit{PMF} \to \mathit{PSDD}$ \\ Compiler};
        \node (px2) at (20,0) {$p(x_2 | \leak)$};
        \draw[->] (vtree_x2) -- (pmfcomp2) node[midway,above] {};
        \draw[->] (px2) -- (pmfcomp2) node[midway,above] {};
        
        \node (psdd_px1) at (24,2) {$\pc(p(x_1 | \leak))$};
        \node (psdd_px2) at (24,-2) {$\pc(p(x_2 | \leak))$};
        
        \draw[->] (pmfcomp1) -- (psdd_px1) node[midway,above] {};
        \draw[->] (pmfcomp2) -- (psdd_px2) node[midway,above] {};
        
\end{tikzpicture}
        }
    }%

    \caption{Given the algorithmic description of \textsc{MixColumn} ($\mathcal{M}$) and the local beliefs $p(v \cbar \leak), v \in \vv$ as inputs to our compilation pipeline, we yield a PSDD representations of all input distributions, denoted $\pc(p(v \cbar \leak))$. Moreover, all PSDDs are pairwise \textit{compatible} for downstream circuit multiplication tasks.}
    \label{fig:compilation}
\end{figure*}

\section{Exact Soft Analytical Side-Channel Attacks}

In this paper, we develop the first exact attack on AES-128 which substantially improves over SASCA and uses far less computational resources than exhaustive enumeration.
Our high-level strategy is to reduce the problem to a tree-shaped factor graph, for which BP becomes an exact inference algorithm \citep{kschischang2001factor,koller2009probabilistic}.
Note that the central problem in Figure~\ref{fig:sasca} (left) is the  sub-graph corresponding to \textsc{MixColumns}, which involves $21$ byte-valued variables and has many loops.
We can now summarize this part into a \emph{single} logical factor 
$\mathcal{M}(\vv)$ as shown in Figure~\ref{fig:sasca} (right), where $\vv$ contains all $21$ variables.
Thus, $\mathcal{M}(\vv)$ is a binary function evaluating to $1$ if and only if $\vv$ corresponds to variable assignment consistent with the entire \textsc{MixColumns} computation.

Running BP in the resulting tree-shaped factor graph yields the exact marginals, by propagating messages from $\mathcal{M}$ to the input bytes $x_i$, computed as \citep{kschischang2001factor},
\begin{align}
\label{eq:message}
    \mu_{\mathcal{M} \to x_i}(x_i) &= \sum_{\vv \setminus x_i} \mathcal{M}(\vv) \cdot \prod_{v_j \in \vv \setminus x_i} p(v_j | \leak)
\end{align}
which requires only two operations, namely \emph{factor multiplication} and \emph{factor summation}.
However, na\"ively computing these messages requires enumerating all $2^{168}$ possible assignments for $\vv$, which is clearly intractable.

In this paper, we therefore build on recent results from knowledge compilation \citep{kcm,sdd} and tractable probabilistic circuits \citep{vergari2020probabilistic,pc_intro,peharz2020einsum}.
The aim of \emph{knowledge compilation} \cite{kcm} is to convert a logical formula, e.g.~given in conjunctive normal form (CNF), into a different target representation that can tractably (i.e.,~in polynomial time) answer certain queries (e.g., the SAT problem). 
In our context, the relevant formula is the boolean factor $\mathcal{M}$ for \textsc{MixColumns}.

Specifically, we compile $\mathcal{M}$ into a \emph{sentential decision diagram} (SDDs) \cite{sdd} which is subsequently converted into its probabilistic extension, \emph{probabilistic SDDs} (PSDDs) \cite{psdd}.
PSDDs are a special case of probabilistic circuit, a structured representation of high-dimensional probability distributions allowing a wide range of tractable probabilistic inference routines \citep{vergari2020probabilistic}. 
Our constructed PSDD represents the uniform distribution over all $\vv$ which are consistent with \textsc{MixColumns}, while inconsistent $\vv$ are assigned probability $0$.

PSDDs allow both factor multiplication and factor summation in polynomial time \citep{vergari2020probabilistic,pc_intro}, as required in \eqref{eq:message}.
Furthermore, they also allow to compute a \emph{most probable evidence} (MPE), i.e.~a probability maximizing assignment.
These tractable routines are the key routines we require for attacking AES.
Before proceeding with implementation details, we review the required background on tractable circuits, in particular SDDs and PSDDs.
At their core, SDDs are hierarchical and structured decompositions of Boolean functions, defined by so-called \emph{compressed partitions}.  

% partitions
\begin{definition}[Compressed Partition \cite{sdd}]
Let $f(\xx, \yy)$ be a Boolean function over disjoint sets of binary variables $\xx$ and $\yy$. 
Any such $f$ can be written as $f = (p_1(\xx) \land s_1(\yy)) \lor \dots \lor (p_k(\xx) \land s_k(\yy))$, where $p_i$ (\textit{primes}) and $s_i$ (\textit{subs}) are Boolean functions over $\xx$ and $\yy$, respectively. Moreover, this decomposition can be constructed such that $p_i \land p_j = \bot$ for $i \neq j$, $p_1 \lor \dots \lor p_k = \top$, and $p_i \neq \bot$ for all $i$. 
Further, we restrict the subs to be distinct, i.e., $s_i \neq s_j$ for all $i \neq j$. 
Then we call $\{(p_1,s_1),\dots,(p_k,s_k)\}$ a \textit{compressed $\xx$-partition} of $f$.
\end{definition}

A structure-defining notion of SDDs are \emph{vtrees}. 
\begin{definition}[Vtree \cite{new_comp_lang}]
    A \textit{vtree}, or \emph{variable tree}, over a set of binary variables $\zz$ is a full, rooted binary tree whose leaves are in a one-to-one correspondence to the variables in $\zz$.
\end{definition}

\begin{definition}[Sentential Decision Diagram \cite{sdd}]
    Let $v$ be a vtree over binary variables $\zz$.
    A \emph{sentential decision diagram} (SDD) is a logical circuit where each internal node is either an \emph{or}-node ($\lor$) or an \emph{and}-node ($\land$), while leaf nodes correspond to variables, their negation, $\top$, or $\bot$. W.l.o.g., we demand that every \emph{or}-node $n$ has $k \geq 1$ $\land$-nodes as children, denoted $n_1,\dots,n_k$. Each $n_i$ has exactly two children, denoted $p_i$, $s_i$. Then, $n$ represents a boolean function $g_n(\xx, \yy)$ (where $\xx, \yy$ are disjoint sets of binary variables) such that $\{(p_i,s_i)\}_{i=1}^k$ is a compressed $\xx$-partition of $g_n$. Further, there exists a node $v'$ in the vtree $v$ such that $\xx = v'_l, \yy = v'_r$ where $v'_l, v'_r$ denote the set of variables mentioned in the left and right subtree of $v'$, respectively. We say that $n$ is \emph{normalized} w.r.t.~$v'$.
    Moreover, we say that an SDD \emph{respects} a vtree $v$ iff every \emph{or}-node $n$ in the SDD is \emph{normalized} w.r.t.~a vtree node in $v$.
\end{definition}

Hence, an SDD is a representation of a Boolean function; due its structural constraints it allows a wide range of logical operations \citep{sdd}.
As mentioned above, we can also think of an SDD as an unnormalized uniform  distribution over all satisfying assignments $\zz$. 
It is straightforward to turn an SDD into a proper probability distribution, a so-called PSDD \cite{psdd}.
 
\begin{definition}[Probabilistic SDD]
Given an SDD, the corresponding PSDD is obtained by replacing all \emph{and}-nodes with \emph{product} nodes and all \emph{or}-nodes with \emph{sum} nodes. 
A \emph{product} node computes the product of its inputs while a \emph{sum} node computes a convex combination of its inputs.
\end{definition}
We denote the PSDD representation of a Boolean function $f$ as $\pc(f)$.
A PSDD is a type of \emph{Probabilistic Circuit} (PC) \cite{pc_intro} with particular properties, allowing 
(i) \emph{arbitrary factor summation (marginalization)} in time linear in the circuit size;
(ii) \emph{circuit multiplication} of two circuits respecting the same vtree, yielding a PSDD whose size is in the worst-case the product of the sizes of the involved circuits;  
(iii) computing a \emph{most-probable-explanation}, i.e.~a probability maximizing variable assignment in time linear in the circuit size.
Hence, PSDDs provide, among others, the operations required for exact message passing (Equation \ref{eq:message}). 
The crucial first step is to compile $\mathcal{M}$.
\subsection{Compiling MixColumn into an SDD}

In order to compile \textsc{MixColumn} (Algorithm \ref{alg:mc}) to an SDD, we represent it as a Boolean function over $168$ binary variables by replacing all variable assignments with equivalence constraints. 
Represented in conjunctive normal form (CNF), $\mathcal{M}$ consists of $648$ clauses with an average of $3.09$ literals within a clause.
To compile the CNF into an SDD representation $\sdd(\mathcal{M})$, we leverage the bottom-up SDD compiler introduced in \cite{dynamic_min_choi}. Initializing the vtree to be left-linear and using dynamic vtree minimization \cite{dynamic_min_choi}, we find that the resulting SDD contains $19k$ sums and products and takes $\approx 30$ seconds to compile on a modern laptop CPU.

Further, to tractably compute the BP message in (\ref{eq:message}), we need to represent the product $\mathcal{M}(\vv) \cdot \prod_{v_j \in \vv \setminus x_i} p(v_j | \leak)$ as a circuit. 
We present two methods for this task, namely 
(i) an ``out-of-the-box'' implementation, compiling the mass functions $p(v_j|\leak)$ upon observing $\leak$ into PSDDs and computing a sequence of circuit products, described in Section \ref{sec:PSDDchain}, and 
(ii) compiling the product symbolically and reducing the problem to \emph{weighted model counting} (WMC) \citep{chavira2008probabilistic}, described in Section \ref{sec:sdd_aux}.
The first techniques makes a slight simplifying sparseness assumption on the variable beliefs, while the second method is exact and can work with arbitrary messages.

\subsection{PSDD Multiplication Chain}
\label{sec:PSDDchain}

First, we convert the SDD representation of \textsc{MixColumn} into a PSDD $\pc(\mathcal{M})$.
Given the vtree of $\pc(\mathcal{M})$ (ranging over $168$ binary variables), we compute a vtree projection \cite{tractable_ops} for each byte $v$ in $\vv$, where each projection ranges over the $8$ bits that make up $v$. It is then straightforward to compile $p(v|\leak)$ into a PSDD that respects the projected vtree. This technique ensures that all pairs of circuits in $\{\pc(\mathcal{M})\} \cup \{\pc(p(v_j|\leak))\}_{j=1}^{21}$ are compatible and can be multiplied on the circuit level.
Thus, the message in Equation \ref{eq:message} can be computed by a tractable marginal query in the circuit product $\pc(\mathcal{M}(\vv)) \cdot \prod_{v_j \in \vv \setminus x_i} \pc(p(v_j | \leak))$. 
In practice, computing this product is infeasible as the circuit size grows in the worst exponentially in the number of multiplied circuits (i.e.~quadratic for two circuits, cubic for three, etc.), and the total number of involved circuits is $22$.

However, when attacking vanilla AES, some of local beliefs $p(v | \leak), v \in \vv$ will typically have \textit{low entropy}, i.e., the probability mass is concentrated on only a few values, often sharing the same hamming weight.
Thus, we can work with \emph{sparse} distributions by setting very small values in the mass functions to $0$: Given $p(v | \leak)$ and a hyperparameter $0 \leq \varepsilon < 1$, we sort the probabilities in ascending order and find the largest $k$ such that the sum of $k$ smallest probabilities is $\leq \varepsilon$. 
We then clamp the $k$ smallest probabilities to $0$ and re-normalize the mass function. 
Empirically, we find that even for values of $\varepsilon$ as small as $10^{-8}$, the runtime of computing this circuit product is comparable to SASCA, while substantially outperforming it.

During compilation, this method can readily leverage sparse probability distributions. 
However, since we fix the vtree after the initial SDD compilation we cannot perform dynamic vtree optimization \cite{dynamic_min_choi} during the PSDD multiplication chain.
Since dynamic compilation is key to achieve small circuits, this out-of-the-box method is computationally infeasible when dropping the low entropy assumption.
This is in particular the case for protection techniques which add noise to the leakage signals.

\subsection{SDD with Auxiliary Byte Indicators}
\label{sec:sdd_aux}

Our second strategy is to perform \emph{symbolic} circuit multiplication, which can take use of dynamic vtree optimization \cite{dynamic_min_choi}.
Specifically, we take the SDD representation of $\sdd(\mathcal{M})$ modelling \emph{bit interactions} and, akin to previous work on SDD compilation \cite{sdd_comp}, successively add auxiliary variables that represent the values of \emph{bytes}. 
For example, let $v$ be a byte that consists of the bits $v^{(1)}, \dots, v^{(8)}$. For each byte variable $v$ in \textsc{MixColumn}, we create $256$ auxiliary boolean variables $b_{v=0},\dots,b_{v=255}$ that indicate the state of $v$. After adding these \emph{byte indicator variables} to the vtree, we conjoin the following constraint to the current SDD:\begin{align*}
&\left( b_{v=0} \Leftrightarrow \left( \neg v^{(1)} \land \dots \land \neg v^{(8)} \right) \right) \land \dots \\ 
&\left(b_{v=255} \Leftrightarrow \left( v^{(1)} \land \dots \land v^{(8)} \right) \right)
\end{align*}
To efficiently encode this into an SDD, we follow the method proposed in \cite{sdd_comp} and compute a base SDD $\alpha$ that represents $\neg b_{v=0} \land \dots \land \neg b_{v=255}$ that we can re-use throughout the compilation.

When compared to the PSDD multiplication chain, the advantage of this approach is that we can utilize dynamic vtree optimization \cite{dynamic_min_choi} \emph{throughout the entire compilation procedure}.
In the SDD with byte indicators, computing the BP messages defined in Equation \ref{eq:message} is effectively a series of weighted model counting (WMC) problems: Let $w(l)$ denote the weight of a literal $l$. To compute $\mu_{\mathcal{M} \to x_i}(x_i = x)$, we set $w(b_{x_i=x}) = 1$, $w(\neg b_{x_i=x}) = 0$, and the weight of all other byte indicators to $w(b_{v=x}) = p(v=x | \leak)$ and $w(\neg b_{v=x}) = 1$ $\forall x \in \{0,\dots,255\}, \forall v \in \vv \setminus x_i$, while we set $w(l) = 1$ for all non-auxiliary literals $l$. Na\"ively, this entails that computing $\{ \mu_{\mathcal{M} \to x_i} \}_{i=1}^4$ requires $4 \cdot 256$ WMC executions. However, it is known that all marginals (i.e.,~messages) can be computed using a \emph{single WMC computation and its derivative} (backward pass) \cite{darwiche2000diff, peharz2015theory}. We thus assume that the number of operations needed to compute all messages is approximately twice the number of operations needed for a bottom-up circuit evaluation.

\subsection{Merging Trick and Conditioning} 
\label{sec:conditioning}

Even when using modern knowledge compilers, constructing the SDD described above is prohibitively expensive. 
Hence, we aid the SDD compiler by performing a number of pre-processing steps. 
First, we apply a known technique in the SCA literature, called the \emph{merging trick} \cite{sasca_coding}: For example, in the SASCA factor graph (Figure \ref{fig:sasca}), $x_{12}$ and $\tilde{x}_{12}$ are connected via a deterministic bijection $f_{\textsc{xtime}}(x_{12}) = \tilde{x}_{12}$. 
Thus, when computing key marginals, there is no need to model both variables in the factor graph---instead, we aggregate the leakage distributions of these variables by updating the local beliefs $p(x_{12}=x | \leak) \gets p(x_{12}=x | \leak) \cdot p(\tilde{x}_{12}=f_{\textsc{xtime}}(x) | \leak)$ for all $x \in \{0,\dots,255\}$. We remove the node $\tilde{x}_{12}$ from the factor graph and omit introducing byte indicator variables for it (or multiplying the PSDD that represents $p(\tilde{x}_{12} | \leak)$). We repeat this trick to remove $\tilde{x}_{23}, \tilde{x}_{34}$, and $\tilde{x}_{41}$. 

Moreover, notice that if we condition on $g$, all \textsc{xor} factors connected to $g$ become indicator functions $\textsc{xor}_g(x, x')$ that evaluate to $1$ iff $x \oplus g = x'$. Thus, we can leverage the merging trick to remove $x'_{12}, x'_{23}, x'_{34}, x'_{41}$ by combining their local beliefs with the distributions of their corresponding left neighbors $\tilde{x}_{12}, \dots, \tilde{x}_{41}$. In the same way, we remove $x_{34}, x_{41}$ by updating the beliefs of $x_{12}$ and $x_{23}$, respectively.
As this simplifies the inference problem drastically, we condition $\sdd(\mathcal{M})$ on a fixed $g \in \{0,\dots,255\}$ and only add byte indicator variables for $10$ bytes in total (corresponding to $\num{2560}$ auxiliary byte indicators). 
In the general case, we can compile a conditional SDD for each value of $g$ (resulting in $256$ SDDs) and easily combine them into a larger SDD by connecting them to an \emph{or}-node.

However, due to the linearity of \textsc{MixColumn}, it suffices to compile a \emph{single} conditional SDD (with arbitrary, fixed $g \in \{0,\dots,255\}$), as we can also run inference queries for different $g' \neq g$ by simply permuting some of the weight functions. This novel technique (1) reduces the time and space complexity of SDD compilation by a factor of $\approx 256$, and (2) opens up the possibility for highly-parallel GPU-centric inference implementations. A more detailed exposition of this technique can be found in Appendix \ref{app:single-circuit}.

Using these techniques, compiling $\sdd(\mathcal{M})$ takes about $7$ hours on a Intel Xeon E5-2670 CPU and the resulting circuit consists of $\approx 20$M nodes and needs $17.4$M sums and $37.2$M products for a bottom-up evaluation. When evaluated $256$ times, this amounts to $6$ times less computations for MPE queries in the corresponding PSDD, and $3.5$ times less computations when computing all marginals.

\subsection{Implementation}

While our approach involves a rather sophisticated compilation and inference pipeline, our implementation makes heavy use of existing software packages for both knowledge compilation and probabilistic inference. Most notably, we leverage \texttt{pysdd} \cite{pysdd}, a Python interface for the \texttt{sdd}\footnote{\url{http://reasoning.cs.ucla.edu/sdd/}} software package, which allows us to easily construct, manipulate and optimize SDDs and their associated vtrees.
This allows us to implement many components of our pipeline (e.g., the ``merging trick'') in a relatively straightforward way.
Our implementation can be found at \url{https://github.com/wedenigt/exsasca}.

\section{Experiments}

To evaluate our method, we use asynchronously captured power traces from an SMT32F415 (ARM) microcontroller \cite{dlsca_defcon} that performs encryption using AES-128. 
The data $\dat = \{(\leak^{(i)}, \kk^{(i)}, \pp^{(i)})\}_{i=1}^n$ contains $n = \num{131072}$ samples, each of which consists of a high-dimensional power trace $\leak \in \mathbb{R}^{\num{20000}}$, and 16 byte key and plaintext vectors $\kk, \pp$, with $512$ unique keys $\kk$. 
For each unique key, the dataset contains $256$ elements, each with a different plaintext $\pp$. 
During the attack phase, we assume the plaintext to be known (\emph{known plaintext attack}).
We use $20\%$ of $\dat$ as a test set $\dattest$, while the remaining $80\%$ of $\dat$ are again split into a training set $\dattrain$ ($82.5\%$ of $\dat \setminus \dattest$) and a validation set $\datval$ ($17.5\%$ of $\dat \setminus \dattest$).
The set of unique keys in $\dattrain, \dattest, \datval$ are non-overlapping, i.e., when validating and testing a side-channel attack, it must reason about keys $\kk$ it has never seen during profiling/training.

\begin{table}[t]
    \caption{Top-1 success rate of different inference methods, measured on the validation dataset $\datval$. $\varepsilon$ controls the level of sparsity in the approximated distributions, where $\varepsilon = 0$ means no approximation (utilizing the method in Section \ref{sec:sdd_aux}).}
    \label{tab:main_res}
    \vskip 0.1in
    \begin{center}
	\begin{tabular}{|l | c | c | c |}
		\hline
    \multicolumn{1}{|l|}{} & \multicolumn{3}{c|}{\textbf{Top-1 Success Rate}} \\
    \cline{2-4}
    \multicolumn{1}{|l|}{\textbf{Inference Method}} & $\varepsilon = 10^{-2}$ & $\varepsilon = 10^{-8}$ & $\varepsilon = 0$\\
	\hline
            Baseline & $0.10 \%$ & $0.10 \%$ & $0.10 \%$ \\ \hline
            SASCA (3 iters)  & $0.52 \%$ & $0.52 \%$ & $0.52 \%$ \\ \hline
            SASCA (50 iters)  & $32.34 \%$ & $33.81 \%$ & $33.76 \%$ \\ \hline
            SASCA (100 iters)  & $34.25 \%$ & $36.04 \%$ & $35.84 \%$ \\ \hline
            ExSASCA + MAR & $57.36 \%$ & $67.37 \%$ & $67.37 \%$ \\ \hline
            ExSASCA + MPE & $\mathbf{57.51} \%$ & $\mathbf{67.60} \%$ & $\mathbf{67.61} \%$  \\ \hline
            
	\end{tabular}
    \end{center}
\end{table}

\subsection{Evaluation}
\label{sec:eval}

As usual in literature, we model only the first round of AES, thus we can independently attack all $4$ branches of \textsc{MixColumn} and reason about the corresponding $4$-byte subkeys independently. 
For a given leakage $\leak$, we utilize the template attack described in \cite{5min} to obtain a set of local beliefs $p(v \cbar \leak)$ for all bytes $v \in \vv$ and for each call to \textsc{MixColumn}.
Using these beliefs, SASCA and ExSASCA + MAR compute marginal posterior distributions $p(k_1 \cbar \leak),\dots,p(k_{16} \cbar \leak)$ and then define the joint key posterior to be $p(\kk | \leak) = \prod_{i=1}^{16} p(k_i | \leak)$. On the other hand, ExSASCA + MPE does \emph{not} make this conditional independence assumption and directly computes $\argmax_{\kk} p(\kk \cbar \leak)$, i.e., the MPE in the \emph{true joint key posterior} according to our probabilistic model $p(\kk \cbar \leak) = p(\kk_{1:4} \cbar \leak) \cdots p(\kk_{13:16} \cbar \leak)$,
where $\kk_{i:j} = (k_i, k_{i+1},\dots, k_{j})$ denotes a subkey.

To simulate \emph{protected} AES implementations, we experiment with corrupting the local belief distributions: For $\alpha \in [0, 1]$ and for all bytes $v \in \vv$, we compute $$\tilde{p}_{\alpha}(v=x \cbar \leak) = (1-\alpha)p(v=x \cbar \leak) + \alpha \frac{1}{256}$$ for all $x \in \{0,\dots,255\}$ and use the set of corrupted beliefs $\tilde{p}_{\alpha}(v \cbar \leak)$ as inputs to different inference methods. In this scenario, we cannot effectively leverage sparse approximations and thus, choose $\varepsilon = 0$ in this experiment.

For evaluation we use \emph{top-1 success rate}: Given a single leakage $\leak$ and the corresponding $16$-byte key $\kk^*$ and plaintext $\pp$, we define an attack to be \emph{successful} if $\kk^* = \argmax_{\kk} p(\kk | \leak)$, i.e., the true key has the highest probability in the joint key posterior. The \emph{top-1 success rate} is the fraction of successful attacks.

\begin{figure}[t]
    \centering
	\includegraphics[width=0.99\linewidth]{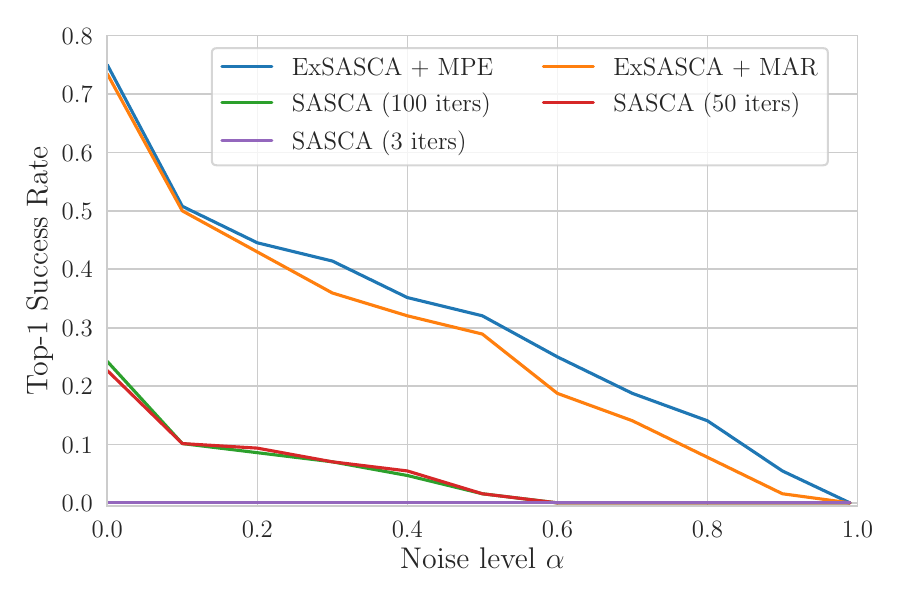}
 \vspace*{-4mm}
    \caption{Top-1 success rate of different inference methods when using corrupted beliefs $\tilde{p}_{\alpha}(v \cbar \leak)$, computed on a batch of $128$ traces in $\datval$ ($\varepsilon = 0$). As $\alpha \to 1$, the local beliefs become increasingly uninformative.}
    \label{fig:noisy_success_rate}
\end{figure}

\subsection{Results}
While SASCA always computes approximate marginals over key bytes, we can use our circuit to compute both (1) exact key marginals (ExSASCA + MAR) and (2) exact most probable evidence assignments directly in the joint key posterior (ExSASCA + MPE). As a baseline, we run BP only in the acyclic factor graph given by the \textsc{AddRoundKey} and \textsc{SubBytes} routines, i.e., we neglect all local beliefs over bytes involved in the computation of \textsc{MixColumn}. 

The results in Table \ref{tab:main_res} show that (1) both ExSASCA variants systematically and substantially outperform all SASCA runs---even when ExSASCA uses sparse approximations and SASCA does not, and (2) directly computing the MPE in the joint key posterior yields a slightly higher success rate. 
Moreover, Figure \ref{fig:noisy_success_rate} demonstrates empirically that as the local beliefs become less informative (e.g.~due to noise), SASCA effectively fails to attack the system while the ExSASCA variants still show success rates of about $20\%$.

\subsection{Computational Complexity}
When $\varepsilon > 0$, the computational cost of ExSASCA depends on both $\varepsilon$ and the entropy of the distributions $p(v \cbar \leak)$. In our experiments, we find that the runtime of ExSASCA with $\varepsilon \geq 10^{-8}$ is still very competitive to a high-performance implementation of SASCA \cite{scalib}: For a single trace, a SASCA takes tens to hundreds of milliseconds on a modern laptop CPU, while our PSDD multiplication chain takes hundreds of milliseconds for $\varepsilon = 10^{-2}$ and seconds for $\varepsilon = 10^{-8}$. In this regime, both SASCA and ExSASCA need three orders of magnitudes less compute than the na\"ive exhaustive inference routine.
Even without approximations ($\varepsilon = 0$), ExSASCA + MPE can perform exact inference with $6$ times less operations than its exhaustive counterpart, while ExSASCA + MAR needs $3.5$ times less operations.

\subsection{Extensions to Larger Inference Problems}

In principle, ExSASCA can also be used to attack more than a single AES round:\footnote{Attacking multiple rounds is also necessary if the key is longer than $128$ bits (e.g., $192$ or $256$ bits).} When neglecting the functional relationship between the key and the so-called ``round keys'', the resulting factor graph is still acyclic (given that \textsc{MixColumns} is represented as a high dimensional factor). However, in this scenario, the usefulness of attacking multiple rounds heavily depends on the beliefs over the round key bytes. Since our dataset does not contain traces for all round key byte values, we cannot learn probabilistic models in the same way as in the single-round attacks and thus, we do not report empirical evaluations of attacking multiple rounds.

\section{Conclusion}

In this work, we introduce ExSASCA, a fast and exact probabilistic inference algorithm for SASCA and demonstrate that ExSASCA substantially outperforms SASCA when attacking the Advanced Encryption Standard (AES).
While SASCA uses loopy belief propagation to compute approximate marginal posteriors over the key bytes, we instead choose to represent the loopy part of the factor graph as a high-dimensional factor, which---conceptually---allows us to perform message passing on a tree.

To efficiently compute messages which involve this factor, we compile it into a tractable probabilistic model that allows us to compute marginals and most probable evidence (MPE) queries in polynomial time of the circuit size. In particular, we develop two distinct approaches for compilation: 
(i) leveraging sparsity in the beliefs about intermediate computations during compilation, we frame the task as a combination of an off-line compilation phase and an on-line sequence of PSDD multiplications. 
(ii) for dense belief distributions, we present a novel dynamic compilation pipeline that produces an SDD that can perform exact inference with $6$ times less operations than exhaustive enumeration. 
We posit that our method opens a new avenue for studying both side-channel attacks and cryptanalysis using the framework of tractable (probabilistic) models and may lead to stronger theoretical guarantees in cryptographic systems.

Future work includes various interesting directions.
In this work we used the hand-designed template attacks proposed by \citep{5min}, but these might be replaced with a deep neural nets in combination with PCs similar as in \citep{stelzner2019faster,tan2019hierarchical,shao2020conditional,spl,gala2024probabilistic}.\footnote{Our preliminary experiments for this paper were actually using deep learning templates, which however under-performed in comparison to \cite{5min}, indicating that designing ``deep templates'' might require some additional engineering effort.}

Moreover, while our central effort in this paper is to detect weaknesses in cryptographic systems, the developed techniques might well be used in other application requiring large-scale integration of probabilistic and logical reasoning, such as system verification and error correcting codes.

\section*{Acknowledgements}

This project has received funding from the European Union's EIC Pathfinder Challenges 2022 programme under grant agreement No 101115317 (NEO). Views and opinions expressed are however those of the author(s) only and do not necessarily reflect those of the European Union or European Innovation Council. Neither the European Union nor the European Innovation Council can be held responsible for them.
\begin{figure}[h]
\centering
\includegraphics[width=0.90\linewidth]{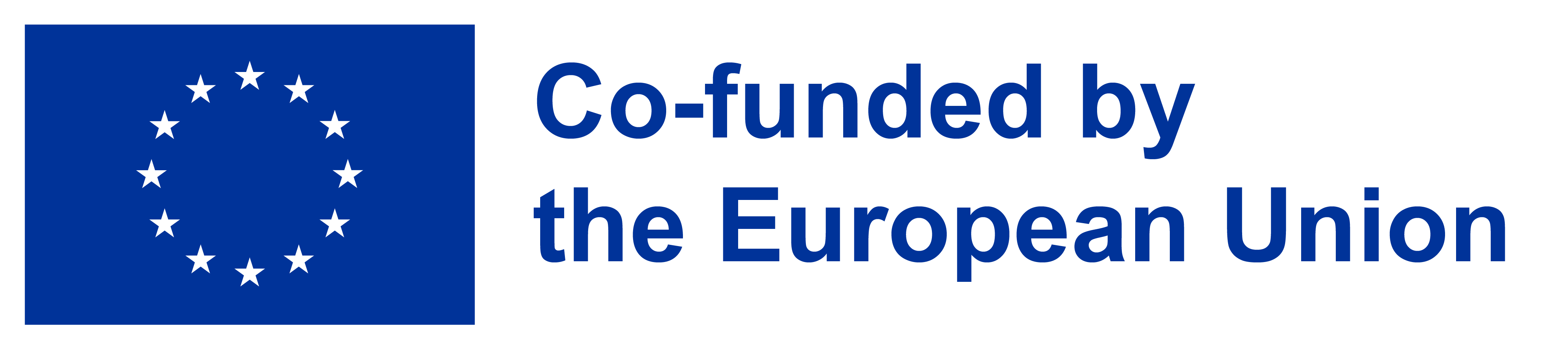}
\end{figure}

\section*{Impact Statement}
This paper presents work whose goals are to advance the field of Machine Learning, specifically the area of exact inference and tractable models, and the field of Cryptography, by providing new tools to analyse weaknesses in cryptographic algorithms via cryptographic attacks.
Hence, our work has a hypothetical potential of dual use. 
However, this potential for dual use is very limited, since (i) we do not provide a new attack, but rather a novel inference method to perform an established attack, and (ii) this established attack is technically challenging.
Moreover, by publishing this improved method we ensure that the scientific and general public is aware of and can react to the connected risks.

% Note use of \abovespace and \belowspace to get reasonable spacing
% above and below tabular lines.


\bibliography{references}

\begin{thebibliography}{37}
\providecommand{\natexlab}[1]{#1}
\providecommand{\url}[1]{\texttt{#1}}
\expandafter\ifx\csname urlstyle\endcsname\relax
  \providecommand{\doi}[1]{doi: #1}\else
  \providecommand{\doi}{doi: \begingroup \urlstyle{rm}\Url}\fi

\bibitem[Ahmed et~al.(2022)Ahmed, Teso, Chang, den Broeck, and Vergari]{spl}
Ahmed, K., Teso, S., Chang, K., den Broeck, G.~V., and Vergari, A.
\newblock Semantic probabilistic layers for neuro-symbolic learning.
\newblock In \emph{NeurIPS}, 2022.

\bibitem[Bronchain et~al.(2021)Bronchain, Cassiers, and Standaert]{5min}
Bronchain, O., Cassiers, G., and Standaert, F.
\newblock Give me 5 minutes: Attacking {ASCAD} with a single side-channel trace.
\newblock \emph{{IACR} Cryptol. ePrint Arch.}, pp.\  817, 2021.

\bibitem[Bursztein \& Picod(2019)Bursztein and Picod]{dlsca_defcon}
Bursztein, E. and Picod, J.-M.
\newblock A hacker guide to deep learning based side channel attacks.
\newblock In CON, D. (ed.), \emph{DEF CON 27}, 2019.

\bibitem[Cassiers \& Bronchain(2023)Cassiers and Bronchain]{scalib}
Cassiers, G. and Bronchain, O.
\newblock Scalib: {A} side-channel analysis library.
\newblock \emph{J. Open Source Softw.}, 8\penalty0 (86):\penalty0 5196, 2023.

\bibitem[Chari et~al.(2002)Chari, Rao, and Rohatgi]{template_attacks}
Chari, S., Rao, J.~R., and Rohatgi, P.
\newblock Template attacks.
\newblock In Jr., B. S.~K., Ko{\c{c}}, {\c{C}}.~K., and Paar, C. (eds.), \emph{Cryptographic Hardware and Embedded Systems - {CHES} 2002, 4th International Workshop, Redwood Shores, CA, USA, August 13-15, 2002, Revised Papers}, volume 2523 of \emph{Lecture Notes in Computer Science}, pp.\  13--28. Springer, 2002.

\bibitem[Chavira \& Darwiche(2008)Chavira and Darwiche]{chavira2008probabilistic}
Chavira, M. and Darwiche, A.
\newblock On probabilistic inference by weighted model counting.
\newblock \emph{Artificial Intelligence}, 172\penalty0 (6-7):\penalty0 772--799, 2008.

\bibitem[Choi \& Darwiche(2013)Choi and Darwiche]{dynamic_min_choi}
Choi, A. and Darwiche, A.
\newblock Dynamic minimization of sentential decision diagrams.
\newblock In desJardins, M. and Littman, M.~L. (eds.), \emph{Proceedings of the Twenty-Seventh {AAAI} Conference on Artificial Intelligence, July 14-18, 2013, Bellevue, Washington, {USA}}, pp.\  187--194. {AAAI} Press, 2013.

\bibitem[Choi et~al.(2013)Choi, Kisa, and Darwiche]{sdd_comp}
Choi, A., Kisa, D., and Darwiche, A.
\newblock Compiling probabilistic graphical models using sentential decision diagrams.
\newblock In van~der Gaag, L.~C. (ed.), \emph{Symbolic and Quantitative Approaches to Reasoning with Uncertainty - 12th European Conference, {ECSQARU} 2013, Utrecht, The Netherlands, July 8-10, 2013. Proceedings}, volume 7958 of \emph{Lecture Notes in Computer Science}, pp.\  121--132. Springer, 2013.

\bibitem[Choi et~al.(2020)Choi, Vergari, and Van~den Broeck]{pc_intro}
Choi, Y., Vergari, A., and Van~den Broeck, G.
\newblock Probabilistic circuits: A unifying framework for tractable probabilistic models.
\newblock 10 2020.

\bibitem[Daemen \& Rijmen(1998)Daemen and Rijmen]{aes}
Daemen, J. and Rijmen, V.
\newblock The block cipher rijndael.
\newblock In Quisquater, J. and Schneier, B. (eds.), \emph{Smart Card Research and Applications, This International Conference, {CARDIS} '98, Louvain-la-Neuve, Belgium, September 14-16, 1998, Proceedings}, volume 1820 of \emph{Lecture Notes in Computer Science}, pp.\  277--284. Springer, 1998.

\bibitem[Darwiche(2000)]{darwiche2000diff}
Darwiche, A.
\newblock A differential approach to inference in bayesian networks.
\newblock In Boutilier, C. and Goldszmidt, M. (eds.), \emph{{UAI} '00: Proceedings of the 16th Conference in Uncertainty in Artificial Intelligence, Stanford University, Stanford, California, USA, June 30 - July 3, 2000}, pp.\  123--132. Morgan Kaufmann, 2000.

\bibitem[Darwiche(2011)]{sdd}
Darwiche, A.
\newblock {SDD:} {A} new canonical representation of propositional knowledge bases.
\newblock In Walsh, T. (ed.), \emph{{IJCAI} 2011, Proceedings of the 22nd International Joint Conference on Artificial Intelligence, Barcelona, Catalonia, Spain, July 16-22, 2011}, pp.\  819--826. {IJCAI/AAAI}, 2011.

\bibitem[Darwiche \& Marquis(2002)Darwiche and Marquis]{kcm}
Darwiche, A. and Marquis, P.
\newblock A knowledge compilation map.
\newblock \emph{J. Artif. Intell. Res.}, 17:\penalty0 229--264, 2002.

\bibitem[de~Colnet \& Mengel(2021)de~Colnet and Mengel]{de2021compilation}
de~Colnet, A. and Mengel, S.
\newblock A compilation of succinctness results for arithmetic circuits.
\newblock \emph{arXiv preprint arXiv:2110.13014}, 2021.

\bibitem[Gala et~al.(2024)Gala, de~Campos, Peharz, Vergari, and Quaeghebeur]{gala2024probabilistic}
Gala, G., de~Campos, C., Peharz, R., Vergari, A., and Quaeghebeur, E.
\newblock Probabilistic integral circuits.
\newblock In \emph{International Conference on Artificial Intelligence and Statistics}, pp.\  2143--2151. PMLR, 2024.

\bibitem[Gandolfi et~al.(2001)Gandolfi, Mourtel, and Olivier]{em_gandolfi}
Gandolfi, K., Mourtel, C., and Olivier, F.
\newblock Electromagnetic analysis: Concrete results.
\newblock In Ko{\c{c}}, {\c{C}}.~K., Naccache, D., and Paar, C. (eds.), \emph{Cryptographic Hardware and Embedded Systems - {CHES} 2001, Third International Workshop, Paris, France, May 14-16, 2001, Proceedings}, volume 2162 of \emph{Lecture Notes in Computer Science}, pp.\  251--261. Springer, 2001.

\bibitem[Guo et~al.(2020)Guo, Grosso, Standaert, and Bronchain]{sasca_coding}
Guo, Q., Grosso, V., Standaert, F., and Bronchain, O.
\newblock Modeling soft analytical side-channel attacks from a coding theory viewpoint.
\newblock \emph{{IACR} Trans. Cryptogr. Hardw. Embed. Syst.}, 2020\penalty0 (4):\penalty0 209--238, 2020.

\bibitem[Kisa et~al.(2014)Kisa, den Broeck, Choi, and Darwiche]{psdd}
Kisa, D., den Broeck, G.~V., Choi, A., and Darwiche, A.
\newblock Probabilistic sentential decision diagrams.
\newblock In Baral, C., Giacomo, G.~D., and Eiter, T. (eds.), \emph{Principles of Knowledge Representation and Reasoning: Proceedings of the Fourteenth International Conference, {KR} 2014, Vienna, Austria, July 20-24, 2014}. {AAAI} Press, 2014.

\bibitem[Knoll(2022)]{bp}
Knoll, C.
\newblock Understanding the behavior of belief propagation.
\newblock \emph{CoRR}, abs/2209.05464, 2022.

\bibitem[Kocher(1996)]{timing_kocher}
Kocher, P.~C.
\newblock Timing attacks on implementations of diffie-hellman, rsa, dss, and other systems.
\newblock In Koblitz, N. (ed.), \emph{Advances in Cryptology - {CRYPTO} '96, 16th Annual International Cryptology Conference, Santa Barbara, California, USA, August 18-22, 1996, Proceedings}, volume 1109 of \emph{Lecture Notes in Computer Science}, pp.\  104--113. Springer, 1996.

\bibitem[Kocher et~al.(1999)Kocher, Jaffe, and Jun]{dpa_kocher}
Kocher, P.~C., Jaffe, J., and Jun, B.
\newblock Differential power analysis.
\newblock In Wiener, M.~J. (ed.), \emph{Advances in Cryptology - {CRYPTO} '99, 19th Annual International Cryptology Conference, Santa Barbara, California, USA, August 15-19, 1999, Proceedings}, volume 1666 of \emph{Lecture Notes in Computer Science}, pp.\  388--397. Springer, 1999.

\bibitem[Koller \& Friedman(2009)Koller and Friedman]{koller2009probabilistic}
Koller, D. and Friedman, N.
\newblock \emph{Probabilistic graphical models: principles and techniques}.
\newblock MIT press, 2009.

\bibitem[Kschischang et~al.(2001)Kschischang, Frey, and Loeliger]{kschischang2001factor}
Kschischang, F.~R., Frey, B.~J., and Loeliger, H.-A.
\newblock Factor graphs and the sum-product algorithm.
\newblock \emph{IEEE Transactions on information theory}, 47\penalty0 (2):\penalty0 498--519, 2001.

\bibitem[MacKay(2003)]{mackay2003information}
MacKay, D.~J.
\newblock \emph{Information theory, inference and learning algorithms}.
\newblock Cambridge university press, 2003.

\bibitem[Meert(2017)]{pysdd}
Meert, W.
\newblock Pysdd.
\newblock In Darwiche, A., Marquis, P., Suciu, D., and Szeider, S. (eds.), \emph{Recent Trends in Knowledge Compilation, Report from Dagstuhl Seminar 17381}, September 2017.
\newblock Dagstuhl Seminar.

\bibitem[Peharz et~al.(2015)Peharz, Tschiatschek, Pernkopf, and Domingos]{peharz2015theory}
Peharz, R., Tschiatschek, S., Pernkopf, F., and Domingos, P.
\newblock {On Theoretical Properties of Sum-Product Networks}.
\newblock In Lebanon, G. and Vishwanathan, S. V.~N. (eds.), \emph{Proceedings of the Eighteenth International Conference on Artificial Intelligence and Statistics}, volume~38 of \emph{Proceedings of Machine Learning Research}, pp.\  744--752, San Diego, California, USA, 09--12 May 2015. PMLR.

\bibitem[Peharz et~al.(2020)Peharz, Lang, Vergari, Stelzner, Molina, Trapp, Van~den Broeck, Kersting, and Ghahramani]{peharz2020einsum}
Peharz, R., Lang, S., Vergari, A., Stelzner, K., Molina, A., Trapp, M., Van~den Broeck, G., Kersting, K., and Ghahramani, Z.
\newblock Einsum networks: Fast and scalable learning of tractable probabilistic circuits.
\newblock In \emph{International Conference on Machine Learning}, pp.\  7563--7574. PMLR, 2020.

\bibitem[Pipatsrisawat \& Darwiche(2008)Pipatsrisawat and Darwiche]{new_comp_lang}
Pipatsrisawat, K. and Darwiche, A.
\newblock New compilation languages based on structured decomposability.
\newblock In Fox, D. and Gomes, C.~P. (eds.), \emph{Proceedings of the Twenty-Third {AAAI} Conference on Artificial Intelligence, {AAAI} 2008, Chicago, Illinois, USA, July 13-17, 2008}, pp.\  517--522. {AAAI} Press, 2008.

\bibitem[Shao et~al.(2020)Shao, Molina, Vergari, Stelzner, Peharz, Liebig, and Kersting]{shao2020conditional}
Shao, X., Molina, A., Vergari, A., Stelzner, K., Peharz, R., Liebig, T., and Kersting, K.
\newblock Conditional sum-product networks: Imposing structure on deep probabilistic architectures.
\newblock In \emph{International Conference on Probabilistic Graphical Models}, pp.\  401--412. PMLR, 2020.

\bibitem[Shen et~al.(2016)Shen, Choi, and Darwiche]{tractable_ops}
Shen, Y., Choi, A., and Darwiche, A.
\newblock Tractable operations for arithmetic circuits of probabilistic models.
\newblock In Lee, D.~D., Sugiyama, M., von Luxburg, U., Guyon, I., and Garnett, R. (eds.), \emph{Advances in Neural Information Processing Systems 29: Annual Conference on Neural Information Processing Systems 2016, December 5-10, 2016, Barcelona, Spain}, pp.\  3936--3944, 2016.

\bibitem[Shen et~al.(2019)Shen, Goyanka, Darwiche, and Choi]{sbn}
Shen, Y., Goyanka, A., Darwiche, A., and Choi, A.
\newblock Structured bayesian networks: From inference to learning with routes.
\newblock In \emph{The Thirty-Third {AAAI} Conference on Artificial Intelligence, {AAAI} 2019, The Thirty-First Innovative Applications of Artificial Intelligence Conference, {IAAI} 2019, The Ninth {AAAI} Symposium on Educational Advances in Artificial Intelligence, {EAAI} 2019, Honolulu, Hawaii, USA, January 27 - February 1, 2019}, pp.\  7957--7965. {AAAI} Press, 2019.

\bibitem[Spreitzer et~al.(2018)Spreitzer, Moonsamy, Korak, and Mangard]{scas}
Spreitzer, R., Moonsamy, V., Korak, T., and Mangard, S.
\newblock Systematic classification of side-channel attacks: {A} case study for mobile devices.
\newblock \emph{{IEEE} Commun. Surv. Tutorials}, 20\penalty0 (1):\penalty0 465--488, 2018.

\bibitem[Standaert(2010)]{intro_sca}
Standaert, F.
\newblock Introduction to side-channel attacks.
\newblock In Verbauwhede, I. M.~R. (ed.), \emph{Secure Integrated Circuits and Systems}, Integrated Circuits and Systems, pp.\  27--42. Springer, 2010.

\bibitem[Stelzner et~al.(2019)Stelzner, Peharz, and Kersting]{stelzner2019faster}
Stelzner, K., Peharz, R., and Kersting, K.
\newblock Faster attend-infer-repeat with tractable probabilistic models.
\newblock In \emph{International Conference on Machine Learning}, pp.\  5966--5975. PMLR, 2019.

\bibitem[Tan \& Peharz(2019)Tan and Peharz]{tan2019hierarchical}
Tan, P.~L. and Peharz, R.
\newblock Hierarchical decompositional mixtures of variational autoencoders.
\newblock In \emph{International Conference on Machine Learning}, pp.\  6115--6124. PMLR, 2019.

\bibitem[Vergari et~al.(2020)Vergari, Choi, Peharz, and Van~den Broeck]{vergari2020probabilistic}
Vergari, A., Choi, Y., Peharz, R., and Van~den Broeck, G.
\newblock Probabilistic circuits: Representations, inference, learning and applications.
\newblock \emph{AAAI Tutorial}, 2020.

\bibitem[Veyrat{-}Charvillon et~al.(2014)Veyrat{-}Charvillon, G{\'{e}}rard, and Standaert]{sasca}
Veyrat{-}Charvillon, N., G{\'{e}}rard, B., and Standaert, F.
\newblock Soft analytical side-channel attacks.
\newblock In Sarkar, P. and Iwata, T. (eds.), \emph{Advances in Cryptology - {ASIACRYPT} 2014 - 20th International Conference on the Theory and Application of Cryptology and Information Security, Kaoshiung, Taiwan, R.O.C., December 7-11, 2014. Proceedings, Part {I}}, volume 8873 of \emph{Lecture Notes in Computer Science}, pp.\  282--296. Springer, 2014.

\end{thebibliography}
\bibliographystyle{icml2024}

%%%%%%%%%%%%%%%%%%%%%%%%%%%%%%%%%%%%%%%%%%%%%%%%%%%%%%%%%%%%%%%%%%%%%%%%%%%%%%%
%%%%%%%%%%%%%%%%%%%%%%%%%%%%%%%%%%%%%%%%%%%%%%%%%%%%%%%%%%%%%%%%%%%%%%%%%%%%%%%
% APPENDIX
%%%%%%%%%%%%%%%%%%%%%%%%%%%%%%%%%%%%%%%%%%%%%%%%%%%%%%%%%%%%%%%%%%%%%%%%%%%%%%%
%%%%%%%%%%%%%%%%%%%%%%%%%%%%%%%%%%%%%%%%%%%%%%%%%%%%%%%%%%%%%%%%%%%%%%%%%%%%%%%
\newpage
\appendix
\onecolumn

\begin{algorithm}[tb]
   \caption{Simplified Beginning of AES-128}
   \label{alg:aes}
\begin{algorithmic}%[1]
   \STATE {\bfseries Input:} Key bytes $k_1,\dots,k_{16}$, Plaintext bytes $p_1,\dots,p_{16}$
   \FOR{$i$ {\bfseries in} $1,\dots,16$}
        \STATE $y_i \gets k_i \oplus p_i$ %\COMMENT{\textsc{AddRoundKey}}
        \STATE $x_i \gets S(y_i)$ %\COMMENT{\textsc{SubBytes}}
   \ENDFOR
   \FOR{$i$ {\bfseries in} $0, 4, 8, 12$}
        \STATE $x_{i+1}^{(m)},\dots,x_{i+4}^{(m)} \gets \textsc{MixColumn}(x_{i+1},\dots,x_{i+4})$ %\Comment{\textsc{MixColumns}}
   \ENDFOR
\end{algorithmic}
\end{algorithm}

\begin{algorithm}[ht]
    \caption{\textsc{MixColumn}}\label{alg:mc}
    \begin{algorithmic}%[1]
    \STATE {\bfseries Input:} Input bytes $x_1,\dots,x_{4}$
    \STATE $x_{12}, \ x_{23}, \ x_{34}, \ x_{41} \gets (x_1 \oplus x_2), (x_2 \oplus x_3), (x_3 \oplus x_4), (x_4 \oplus x_1)$
    \STATE $g \gets x_{12} \oplus x_{34}$

    \STATE $\tilde{x}_{12}, \ \tilde{x}_{23}, \ \tilde{x}_{34}, \ \tilde{x}_{41} \gets \textsc{xtime}(x_{12}), \textsc{xtime}(x_{23}), \textsc{xtime}(x_{34}), \textsc{xtime}(x_{41})$

    \STATE $x'_{12}, \ x'_{23}, \ x'_{34}, \ x'_{41} \gets (\tilde{x}_{12} \oplus g), (\tilde{x}_{23} \oplus g), (\tilde{x}_{34} \oplus g), (\tilde{x}_{41} \oplus g)$

    \STATE $x^{(m)}_{1}, \ x^{(m)}_{2}, \ x^{(m)}_{3}, \ x^{(m)}_{4} \gets (x_1 \oplus x'_{12}), (x_2 \oplus x'_{23}), (x_3 \oplus x'_{34}), (x_4 \oplus x'_{41})$
    \STATE {\bfseries Return:} $x^{(m)}_{1}, \ x^{(m)}_{2}, \ x^{(m)}_{3}, \ x^{(m)}_{4}$
    \end{algorithmic}
\end{algorithm}

\section{Single-Circuit Compilation}
\label{app:single-circuit}
As detailed in Section \ref{sec:conditioning}, we compile $\mathcal{M}$ conditioned on some arbitrary, but fixed $g \in \{0,\dots,255\}$, denoted $\sdd(\mathcal{M} \cbar g)$. 
Consider the set of models $\mathcal{X}_g = \{\vv \ \text{ s.t. } (\mathcal{M} \cbar g)(\vv) = 1 \}$. Since $g = x_1 \oplus x_2 \oplus x_3 \oplus x_4$, it is easy to see that $g$ partitions the set of \emph{all} models into equally sized sets, i.e., $|\mathcal{X}_g| = |\mathcal{X}_{g'}|$ for all byte-valued $g, g'$.
Thus, there exists a \emph{bijection} $\phi_{g \to g'}: \mathcal{X}_g \to \mathcal{X}_{g'}$ that maps a model from $\mathcal{X}_g$ to a model in $\mathcal{X}_{g'}$. Importantly, due to the linear structure of \textsc{MixColumn}, we can define such a bijection by \emph{independently} mapping the individual bytes $v \in \vv$:
\begin{equation}
\left( \phi_{g \to g'}(\vv) \right)_i = \phi^{v_i}_{g \to g'}(v_i)
\end{equation}
Let $\vv_g \in \mathcal{X}_g$. Since we can use the merging trick detailed in Section \ref{sec:conditioning}, it suffices to consider the subset of bytes in $\vv_g$, namely $\hat{\vv}_g = (x_1, x_2, x_3, x_4, x_{12}, x_{23}, x_1^{(m)}, x_2^{(m)}, x_3^{(m)}, x_4^{(m)})^{\top}$. To map this to some $\hat{\vv}_{g'}$, we set 
\begin{itemize}
    \item $\phi^{x_4}_{g \to g'}(x_4) = x_4 \oplus g \oplus g'$
    \item $\phi^{x_1^{(m)}}_{g \to g'}(x_1^{(m)}) = x_1^{(m)} \oplus g \oplus g'$
    \item $\phi^{x_2^{(m)}}_{g \to g'}(x_2^{(m)}) = x_2^{(m)} \oplus g \oplus g'$
    \item $\phi^{x_3^{(m)}}_{g \to g'}(x_3^{(m)}) = x_3^{(m)} \oplus f_{\textsc{xtime}}(g \oplus g')$
    \item $\phi^{x_4^{(m)}}_{g \to g'}(x_3^{(m)}) = x_3^{(m)} \oplus f_{\textsc{xtime}}(g \oplus g') \oplus g \oplus g'$
\end{itemize}
and $\phi^{v}_{g \to g'}(v) = v$ for all remaining bytes $v \in \hat{\vv}_g$. Consequently, we can exploit this fact to compute a weighted model count over $\mathcal{X}_{g'}$ using $\sdd(\mathcal{M} \cbar g)$ with $g' \neq g$ by (1) performing the merge trick with $g'$ instead of $g$, and (2) by constructing a new weight function $w'$ that first applies the corresponding byte-wise bijection before invoking the original weight function $w$, i.e., $w'(v) = w\left(\phi^{v}_{g \to g'}(v) \right)$. Thus, we keep the circuit $\sdd(\mathcal{M} \cbar g)$ fixed and compute the weighted model count $256$ times, where each computation is performed with a different weight function.
%%%%%%%%%%%%%%%%%%%%%%%%%%%%%%%%%%%%%%%%%%%%%%%%%%%%%%%%%%%%%%%%%%%%%%%%%%%%%%%
%%%%%%%%%%%%%%%%%%%%%%%%%%%%%%%%%%%%%%%%%%%%%%%%%%%%%%%%%%%%%%%%%%%%%%%%%%%%%%%

\end{document}